\documentclass[11pt,letterpaper]{article}
\usepackage{emnlp2016}
\usepackage{times}
\usepackage{latexsym}
\usepackage{graphicx}
\usepackage{amsmath}
\usepackage[linesnumbered,noend]{algorithm2e}
\usepackage{url}

\emnlpfinalcopy

\def\emnlppaperid{***}

\title{AMR-to-text generation as a Traveling Salesman Problem}

\author{Linfeng Song$^1$, Yue Zhang$^3$, Xiaochang Peng$^1$, Zhiguo Wang$^2$ \and Daniel Gildea$^1$ \\
        $^1$Department of Computer Science, University of Rochester, Rochester, NY 14627 \\ 
        $^2$IBM T.J. Watson Research Center, Yorktown Heights, NY 10598  \\
        $^3$Singapore University of Technology and Design}

\date{}

\begin{document}

\maketitle

\begin{abstract}
  The task of AMR-to-text generation is to generate grammatical text that sustains the semantic meaning for a given AMR graph.
  We attack the task by 
  first partitioning the AMR graph into smaller fragments, and then generating the translation for each fragment, before finally deciding the order by solving an asymmetric generalized traveling salesman problem (AGTSP).
  A Maximum Entropy classifier is trained to estimate the traveling costs, and a TSP solver is used to find the optimized solution.
  The final model reports a BLEU score of 22.44 on the SemEval-2016 Task8 dataset.
\end{abstract}

\section{Introduction}

\begin{figure}
\centering
\includegraphics[scale=.6]{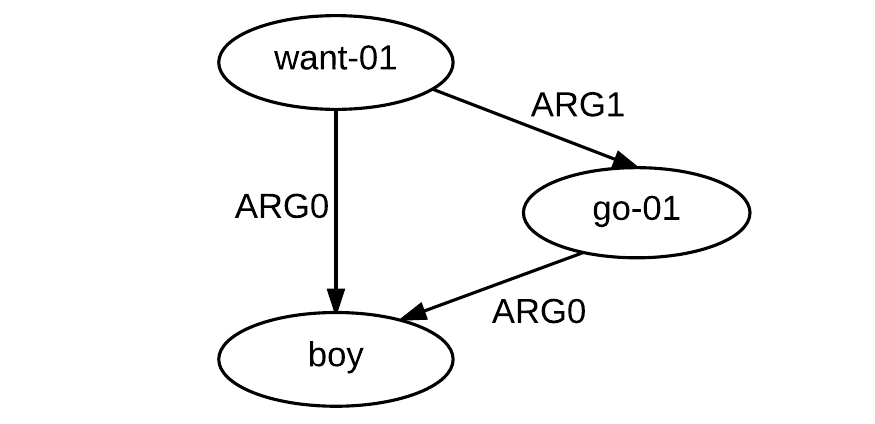}
\caption{AMR graph for ``The boy wants to go''.}
\label{fig:amr}
\end{figure}

Abstract Meaning Representation (AMR) \cite{banarescu-EtAl:2013:LAW7-ID} is a semantic formalism encoding the meaning of a sentence as a rooted, directed graph.
Shown in Figure \ref{fig:amr}, the nodes of an AMR graph (e.g.\ ``boy'', ``go-01'' and ``want-01'') represent concepts, and the edges (e.g.\ ``ARG0'' and ``ARG1'') represent relations between concepts. 
AMR jointly encodes a set of different semantic phenomena, which makes it useful in applications like question answering and semantics-based machine translation.
AMR has served as an intermediate representation for various text-to-text NLP applications, such as statistical machine translation
(SMT) \cite{jones2012semantics}.

The task of AMR-to-text generation is to generate grammatical text containing the same semantic meaning as a given AMR graph.
This task is important yet also challenging since each AMR graph usually has multiple corresponding sentences, and syntactic structure and function words are abstracted away when transforming a sentence into AMR \cite{banarescu-EtAl:2013:LAW7-ID}.
There has been work dealing with text-to-AMR parsing
\cite{flanigan2014discriminative,wang-xue-pradhan:2015:NAACL-HLT,peng2015synchronous,vanderwende2015amr,pust2015parsing,artzi-lee-zettlemoyer:2015:EMNLP}.
On the other hand, relatively little work has been done on AMR-to-text generation.
One recent exception is \newcite{jeff2016amrgen}, who first generate a spanning tree for the input AMR graph, and then apply a tree transducer to generate the sentence.
Here, we directly generate the sentence from an input AMR by treating AMR-to-text generation as a variant of the traveling salesman problem (TSP).

Given an AMR as input, our method first cuts the graph into several rooted and connected fragments (sub-graphs), and then finds the translation for each fragment, before finally generating the sentence for the whole AMR by ordering the translations.
To cut the AMR and translate each fragment, we match the input AMR with rules, each consisting of a rooted, connected AMR fragment and a corresponding translation.
These rules serve in a similar way to rules in SMT models.
We learn the rules by a modified version of the sampling algorithm of \newcite{peng2015synchronous}, and use the rule matching algorithm of \newcite{cai2013smatch}.

For decoding the fragments and synthesizing the output, we define a \emph{cut} to be a subset of matched rules without overlap that covers the AMR, and an \emph{ordered cut} to be a cut with the rules being ordered.
To generate a sentence for the whole AMR, we search for an ordered cut, and concatenate translations of all rules in the cut.
TSP is used to traverse different cuts and determine the best order.
Intuitively, our method is similar to phrase-based SMT, which first cuts the input sentence into phrases, then obtains the translation for each source phrase, before finally generating the target sentence by ordering the translations.
Although the computational cost of our method is low, the initial experiment is promising, yielding a BLEU score of 22.44 on a standard benchmark.

\section{Method}

We reformulate the problem of AMR-to-text generation as an asymmetric generalized traveling salesman problem (AGTSP), a variant of TSP.

\subsection{TSP and its variants}

Given a \emph{non-directed} graph $G_N$ with $n$ cities, supposing that there is a traveling cost between each pair of cities, TSP tries to find a tour of the minimal total cost visiting each city exactly once. 
In contrast, the asymmetric traveling salesman problem (ATSP) tries to find a tour of the minimal total cost on a \emph{directed} graph, where the traveling costs between two nodes are different in each direction.
Given a directed graph $G_D$ with $n$ nodes, which are clustered into $m$ groups, the asymmetric generalized traveling salesman problem (AGTSP) tries to find a tour of the minimal total cost visiting each \emph{group} exactly once.

\subsection{AMR-to-text Generation as AGTSP}
\label{sec:agtsp}

Given an input AMR $A$, each node in the AGTSP graph can be represented as $(c,r)$, where $c$ is a concept in $A$ and $r = (A_{sub},T_{sub})$ is a rule that consists of an AMR fragment containing $c$ and a translation of the fragment.
We put all nodes containing the same concept into one group, thereby translating each concept in the AMR exactly once.

\begin{figure}
\centering
\includegraphics[scale=.7]{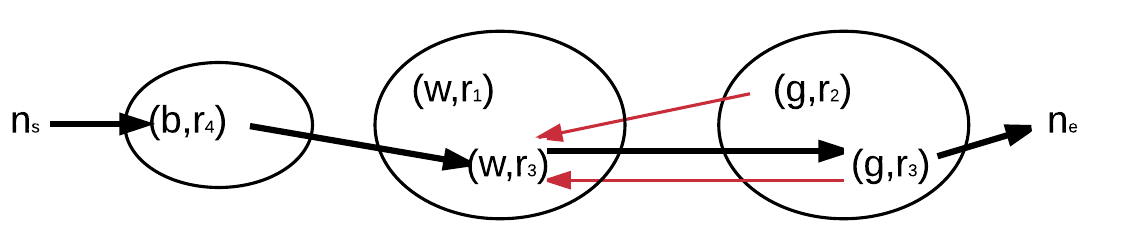}
\caption{An example AGTSP graph}
\label{fig:agtsp}
\end{figure}

To show a brief example, consider the AMR in Figure \ref{fig:amr} and the following rules, 

\vspace{2mm}

\noindent
\begin{tabular}{ll} 
\hline
  $r_1$ & (w/want-01) $|||$ wants \\
  $r_2$ & (g/go-01) $|||$ to go \\  
  $r_3$ & (w/want-01 :ARG1 g/go-01) $|||$ wants to go \\
  $r_4$ & (b/boy) $|||$ The boy \\
\hline
\end{tabular}

~~

\noindent
We build an AGTSP graph in Figure \ref{fig:agtsp}, where each circle represents a group and each tuple (such as $(b,r_4)$) represents a node in the AGTSP graph.
We add two nodes $n_s$ and $n_e$ representing the start and end nodes respectively.
Each belongs to a specific group that only contains that node, and a tour always starts with $n_s$ and ends with $n_e$.
Legal moves are shown in black arrows, while illegal moves are shown in red.
One legal tour is $n_s \rightarrow (b, r_4) \rightarrow (w, r_3) \rightarrow (g, r_3) \rightarrow n_e$.
The order in which nodes within a rule are visited is arbitrary; 
for a rule with $N$ concepts, the number of visiting orders is $O(N!)$.
To reduce the search space, we enforce the breadth first order by setting costs to zero or infinity.
In our example, the traveling cost from $(w, r_3)$ to $(g, r_3)$ is 0, 
while the traveling cost from $(g, r_3)$ to $(w, r_3)$ is infinity.
Traveling from $(g,r_2)$ to $(w,r_3)$ also has infinite
cost, since there is overlap on the concept ``w/want-01'' between them.

\begin{algorithm}[t] 
 \KwData{Nodes in AGTSP graph $G$}
 \KwResult{Traveling Cost Matrix $T$}
 $n_s \leftarrow ($``\textless s\textgreater'',``\textless s\textgreater'' $)$\;
 $n_e \leftarrow ($``\textless /s\textgreater'',``\textless /s\textgreater'' $)$\;
 T[$n_s$][$n_e$] $\leftarrow$ $\infty$\;
 T[$n_e$][$n_s$] $\leftarrow$ 0\;
 \For{$n_i \leftarrow (c_i,r_i)$ \textbf{in} $G$}{
  \eIf{$c_i$ = $r_i$.frag.first}{
   T[$n_s$][$n_i$] $\leftarrow$ ModelScore($n_s$,$n_i$)\;
   }{
   T[$n_s$][$n_i$] $\leftarrow$ $\infty$;
  }
  \eIf{$c_i$ = $r_i$.frag.last}{
  T[$n_i$][$n_e$] $\leftarrow$ ModelScore($n_i$,$n_e$)\;
  }{
  T[$n_i$][$n_e$] $\leftarrow$ $\infty$;
  }
 }
 \For{$n_i \leftarrow (c_i,r_i)$ \textbf{in} $G$}{
 	\For{$n_j \leftarrow (c_j,r_j)$ \textbf{in} $G$}{
       \uIf{$r_i$ = $r_j$ \textrm{and} $r_i$.frag.next($c_i$) = $c_j$}{
       	  T[$n_i$][$n_j$] $\leftarrow$ 0
	   }\uElseIf{$r_i$.frag $\cap$ $r_j$.frag = $\emptyset$ \textrm{and} $c_i$ = $r_i$.frag.last \textrm{and} $c_j$ = $r_j$.frag.first}{
          T[$n_i$][$n_j$] $\leftarrow$ ModelScore($n_i$,$n_j$)
       }
       \uElse{
       	  T[$n_i$][$n_j$] $\leftarrow$ $\infty$
       }
    }
 }
 \caption{Traveling cost algorithm}
 \label{algo:travel_cost}
\end{algorithm}
The traveling cost is calculated by Algorithm \ref{algo:travel_cost}.
We first add $n_s$ and $n_e$ serving the same function as Figure \ref{fig:agtsp}.
The traveling cost from $n_s$ directly to $n_e$ is infinite, since a tour has to go through other nodes before going to the end.
On the other hand, the traveling cost from $n_e$ to $n_s$ is 0 (Lines 3-4), 
as a tour always goes back to the start after reaching the end.
The traveling cost from $n_s$ to $n_i=(c_i,r_i)$ is the model score only if $c_i$ is the first node of the AMR fragment of $r_i$, otherwise the traveling cost is infinite (Lines 6-9).
Similarly, the traveling cost from $n_i$ to $n_e$ is the model score only if $c_i$ is the last node of the fragment of $r_i$. 
Otherwise, it is infinite (Lines 10-13).
The traveling cost from $n_i=(c_i,r_i)$ to $n_j=(c_j,r_j)$ is 0 if $r_i$ and $r_j$ are the same rule and $c_j$ is the next node of $c_i$ in the AMR fragment of $r_i$ (Lines 16-17).

A tour has to travel through an AMR fragment before jumping to another fragment.
We choose the breadth-first order of nodes within the same rule, 
which is guaranteed to exist, as each AMR fragment is rooted and connected.
Costs along the breadth-first order within a rule $r_i$ are set to 0, while other costs
with a rule are infinite.

If $r_i$ is not equal to $r_j$, then the traveling cost is the model score if there is no overlap between $r_i$ and $r_j$'s AMR fragment and it moves from $r_i$'s last node to $r_j$'s first node (Lines 18-19), otherwise the traveling cost is infinite (Lines 20-21).
All other cases are illegal and we assign infinite traveling cost.
We do not allow traveling between overlapping nodes, whose AMR fragments share common concepts.
Otherwise the traveling cost is evaluated by a maximum entropy model, which will be discussed in detail in Section \ref{sec:travel_cost}.

\subsection{Rule Acquisition}
\label{sec:rule_acq}

We extract rules from a corpus of (sentence, AMR) pairs using the method of \newcite{peng2015synchronous}.
Given an aligned (sentence, AMR) pair, a \emph{phrase-fragment pair} is a pair $([i,j],f)$, where $[i,j]$ is a span of the sentence and $f$ represents a connected and rooted AMR fragment.
A \emph{fragment decomposition forest} consists of all possible phrase-fragment pairs that satisfy the alignment agreement for phrase-based MT \cite{koehn2003statistical}.
The rules that we use for generation are the result of applying 
an MCMC procedure
to learn a set of likely phrase-fragment pairs from the forests 
containing all possible pairs.
One difference from the work of \newcite{peng2015synchronous} 
is that, while they require the string side to be tight (does not include unaligned words on both sides),  
we expand the tight phrases to incorporate unaligned words on both sides.
The intuition is that they do text-to-AMR parsing, 
which often involves discarding function words,
while our task is AMR-to-text generation, 
and we need to be able to fill in these unaligned words.
Since incorporating unaligned words will introduce noise, we rank the translation candidates for each AMR fragment by their counts in the training data, and select the top $N$ candidates.%
\footnote{Our code for grammar induction can be downloaded from
  https://github.com/xiaochang13/AMR-generation}

We also generate \emph{concept rules} which directly use a morphological string of the concept for translation. 
For example, for concept ``w/want-01'' in Figure \ref{fig:amr}, we generate concept rules such as ``(w/want-01) $|||$ want'', ``(w/want-01) $|||$ wants'', ``(w/want-01) $|||$ wanted'' and ``(w/want-01) $|||$ wanting''.
The algorithm (described in section \ref{sec:agtsp}) will choose the most suitable one from the rule set.
It is similar to most MT systems in creating a translation candidate for each word, besides normal translation rules.
It is easy to guarantee that the rule set can fully cover every input AMR graph.

Some concepts (such as ``have-rel-role-91'') in an AMR graph do not contribute to the final translation,
and we skip them when generating concept rules.
Besides that, we use a verbalization list\footnote{http://amr.isi.edu/download/lists/verbalization-list-v1.06.txt} for concept rule generation.
For rule ``VERBALIZE peacekeeping TO keep-01 :ARG1 peace'', we will create a concept rule ``(k/keep-01 :ARG1 (p/peace)) $|||$ peacekeeping'' if the left-hand-side fragment appears in the target graph.

\subsection{Traveling cost}
\label{sec:travel_cost}

Considering an AGTSP graph whose nodes are clustered into $m$ groups, we define the traveling cost for a tour $T$ in Equation \ref{eq:tc}:
\begin{equation}
\label{eq:tc}
  \textrm{cost}(n_s,n_e) = -\sum_{i=0}^{m}\log p(\textrm{``yes''}|n_{T_i},n_{T_{i+1}})
\end{equation}
where $n_{T_0}=n_s$, $n_{T_{m+1}}=n_e$ and each $n_{T_i}$ ($i \in [1\ldots m]$) belongs to a group that is different from all others.  Here
$p(\textrm{``yes''}|n_j,n_i)$ represents a 
learned score for a move from $n_j$ to $n_i$.
The choices before $n_{T_i}$ are independent from choosing $n_{T_{i+1}}$ given $n_{T_i}$ because of the Markovian property of the TSP problem.
Previous methods \cite{zaslavskiy2009phrase} evaluate traveling costs $p(n_{T_{i+1}}|n_{T_i})$ by using a language model.
Inevitably some rules may only cover one translation word, making only bigram language models naturally applicable.
\newcite{zaslavskiy2009phrase} introduces a method for incorporating a trigram language model.
However, as a result, the number of nodes in the AGTSP graph grows exponentially.

To tackle the problem, 
we treat it as a \emph{local} binary (``yes'' or ``no'') classification problem whether we should move to $n_j$ from $n_i$.
We train a maximum entropy model, where $p(\textrm{``yes''}|n_i,n_j)$ is defined as:
\begin{multline}
\label{eq:maxent}
  p(\textrm{``yes''}|n_i,n_j) = \\
  \frac{1}{Z(n_i,n_j)} \textrm{exp} \Big[ \sum_{i=1}^k \lambda_i f_i(\textrm{``yes''},n_i,n_j) \Big]
\end{multline}
The model uses 3 real-valued features: a language model score, the word count of the concatenated translation from $n_i$ to $n_j$,
and the length of the shortest path from $n_i$'s root to $n_j$'s root in the input AMR.
If either $n_i$ or $n_j$ is the start or end node, we set the path length to 0.
Using this model, we can use whatever N-gram we have at each time.
Although language models favor shorter translations, word count will balance the effect, which is similar to MT systems.
The length of the shortest path is used as a feature because the concepts whose translations are adjacent usually have lower path length than others.

\section{Experiments}
\label{sec:exp}

\subsection{Setup}

We use the dataset of SemEval-2016 Task8 (Meaning Representation Parsing), which contains 16833 training instances, 1368 dev instances and 1371 test instances.
Each instance consists of an AMR graph and a sentence representing the same meaning.
Rules are extracted from the training data, and hyperparameters are tuned on the dev set.
For tuning and testing, we filter out sentences that have more than 30 words, resulting in 1103 dev instances and 1055 test instances.
We train a 4-gram language model (LM) with gigaword (LDC2011T07), and use BLEU \cite{papineni2002bleu} as the evaluation metric.
To solve the AGTSP, we use Or-tool\footnote{https://developers.google.com/optimization/}.

Our graph-to-string rules are reminiscent of phrase-to-string rules in phrase-based MT (PBMT).
We compare our system to a baseline (\emph{PBMT}) that first linearizes the input AMR graph by breadth first traversal, and then adopts the PBMT system from Moses\footnote{http://www.statmt.org/moses/} to translate the linearized AMR into a sentence. 
To traverse the children of an AMR concept, we use the original order in the text file. 
The MT system is trained with the default setting on the same dataset and LM.
We also compare with \emph{JAMR-gen}\footnote{https://github.com/jflanigan/jamr/tree/Generator} \cite{jeff2016amrgen}, 
which is trained on the same dataset but with a 5-gram LM from gigaword (LDC2011T07).

To evaluate the importance of each module in our system, we develop the following baselines: 
\emph{OnlyConceptRule} uses only the concept rules,
\emph{OnlyInducedRule} uses only the rules induced from the fragment decomposition forest,
\emph{OnlyBigramLM} uses both types of rules, but the traveling cost is evaluated by a bigram LM trained with gigaword.

\begin{table}[t]
\centering
 \begin{tabular}{|l|l|l|} 
 \hline
 System & Dev & Test \\
 \hline
 \hline
 PBMT & 13.13 & 16.94 \\
 \hline
 OnlyConceptRule & 13.15 & 14.93 \\
 OnlyInducedRule & 17.68 & 18.09 \\
 OnlyBigramLM    & 17.19 & 17.75 \\
 All             & 21.12 & 22.44 \\
 \hline
 JAMR-gen & \textbf{23.00} & \textbf{23.00} \\
 \hline
 \end{tabular}
 \caption{Main results.} \label{tab:rst}
\end{table}

\subsection{Results}

The results are shown in Table \ref{tab:rst}. Our method (\emph{All}) significantly outperforms the baseline (\emph{PBMT}) on both the dev and test sets.
\emph{PBMT} does not outperform \emph{OnlyBigramLM} and \emph{OnlyInducedRule}, demonstrating that our rule induction algorithm is effective.
We consider rooted and connected fragments from the AMR graph, and the TSP solver finds better solutions than beam search, as consistent with \newcite{zaslavskiy2009phrase}.
In addition, \emph{OnlyInducedRule} is significantly better than \emph{OnlyConceptRule}, showing the importance of induced rules on performance.
This also confirms the reason that \emph{All} outperforms \emph{PBMT}.
This result confirms our expectation that concept rules, which are used for fulfilling the coverage of an input AMR graph in case of OOV, 
are generally not of high quality.
Moreover, \emph{All} outperforms \emph{OnlyBigramLM} showing that our maximum entropy model is stronger than a bigram language model. 
Finally, \emph{JAMR-gen} outperforms \emph{All},
while \emph{JAMR-gen} uses a higher order language model than \emph{All} (5-gram VS 4-gram).

For rule coverage, around 31\% AMR graphs and 84\% concepts in the development set are covered by our induced rules extracted from the training set.


\subsection{Analysis and Discussions}

\begin{table}
\centering
 \begin{tabular}{|l|} 
 \hline
 (w / want-01 \\
 ~~~~~~~~:ARG0 (b / boy) \\
 ~~~~~~~~:ARG1 (b2 / believe-01 \\
 ~~~~~~~~~~~~~~~~:ARG0 (g / girl) \\
 ~~~~~~~~~~~~~~~~:ARG1 b)) \\
 \hline
 \textbf{Ref:}~~the boy wants the girl to believe him \\
 \hline
 \textbf{All:}~~a girl wanted to believe him \\
 \hline
 \textbf{JAMR-gen:}~~boys want the girl to believe \\
 \hline
 \end{tabular}
 \caption{Case study.} \label{tab:case}
\end{table}

We further analyze \emph{All} and \emph{JAMR-gen} with an example AMR and show the AMR graph, the reference, and results in Table \ref{tab:case}.
First of all, both \emph{All} and \emph{JAMR-gen} outputs a reasonable translation containing most of the meaning from the AMR.  
On the other hand, \emph{All} fails to recognize ``boy'' as the subject.
The reason is that the feature set does not include edge labels, such as ``ARG0'' and ``ARG1''.
Finally, neither \emph{All} and \emph{JAMR-gen} can handle the situation when a re-entrance node (such as ``b/boy'' in example graph of Table \ref{tab:case}) need to be translated twice. 
This limitation exists for both works.

\section{Related Work}

Our work is related to prior work on AMR \cite{banarescu-EtAl:2013:LAW7-ID}.
There has been a list of work on AMR parsing 
\cite{flanigan2014discriminative,wang-xue-pradhan:2015:NAACL-HLT,peng2015synchronous,vanderwende2015amr,pust2015parsing,artzi-lee-zettlemoyer:2015:EMNLP}, 
which predicts the AMR structures for a given sentence.
On the reverse direction, \newcite{jeff2016amrgen} and our work here study sentence generation from a given AMR graph.
Different from \newcite{jeff2016amrgen} who map a input AMR graph into a tree before linearization,
we apply synchronous rules consisting of AMR
graph fragments and text to directly transfer a AMR graph into a sentence.
In addition to AMR parsing and generation, there has also been work using AMR as a semantic representation in machine translation \cite{jones2012semantics}.

Our work also belongs to the task of text generation \cite{reiter1997building}.
There has been work on generating natural language text from a bag of words \cite{wan-EtAl:2009:EACL,zhang2015discriminative}, 
surface syntactic trees \cite{zhang2013partial,song2014joint},
deep semantic graphs \cite{bohnet-EtAl:2010:PAPERS} 
and logical forms \cite{white2004reining,white-rajkumar:2009:EMNLP}.
We are among the first to investigate generation from AMR, which is a different type of semantic representation.

\section{Conclusion}

In conclusion, we showed that a TSP solver with a few real-valued features can be useful for AMR-to-text generation.
Our method is based on a set of graph to string rules, yet significantly better than a PBMT-based baseline.
This shows that our rule induction algorithm is effective and that the TSP solver finds better solutions than beam search.

\section*{Acknowledgments}

We are grateful for the help of Jeffrey Flanigan, Lin Zhao, and Yifan He. 
This work was funded by NSF IIS-1446996, and a Google Faculty Research Award.
Yue Zhang is funded by NSFC61572245 and T2MOE201301 from Singapore Ministry of Education.

\bibliography{emnlp2016}
\bibliographystyle{emnlp2016}

\end{document}